\documentclass[lettersize,journal]{IEEEtran}
\usepackage{amsmath,amsfonts}
\usepackage{algorithmic}
\usepackage{algorithm}
\usepackage{array}
\usepackage[caption=false,font=normalsize,labelfont=sf,textfont=sf]{subfig}
\usepackage{textcomp}
\usepackage{stfloats}
\usepackage{url}
\usepackage{verbatim}
\usepackage{graphicx}
\usepackage{cite}
\usepackage[normalem]{ulem}
\useunder{\uline}{\ul}{}
\usepackage{booktabs}
\usepackage{float} 
\usepackage{tabularx}

\begin{document}

\title{Prompt Engineering for Healthcare: Methodologies and Applications}
\author{Jiaqi Wang$^*$, Enze Shi$^*$\thanks{$^*$Co-first author}, Sigang Yu, Zihao Wu, Chong Ma, Haixing Dai, Qiushi Yang, Yanqing Kang, Jinru Wu, Huawen Hu, Chenxi Yue, Haiyang Zhang, Yiheng Liu, Yi Pan, Zhengliang Liu, Lichao Sun, \\Xiang Li, Bao Ge, Xi Jiang, Dajiang Zhu, Yixuan Yuan, Dinggang Shen, Tianming Liu, and Shu Zhang
\thanks{Corresponding author: Shu Zhang}
\thanks{Jiaqi Wang, Enze Shi, Sigang Yu, Yanqing Kang, Jinru Wu, Huawen Hu, Chenxi Yue, Haiyang Zhang, Shu Zhang are with the School of Computer Science, Northwestern Polytechnical University, Xi'an 710072, China. Chong Ma is with the School of Automation, Northwestern Polytechnical University, Xi'an 710072, China. (e-mail: \{jiaqi.wang, ezshi, sgyu, yanqing.kang, jinru.wu, huawenhu, chenxi.yue, haiyang.zhang\}@mail.nwpu.edu.cn; shu.zhang@nwpu.edu.cn; mc-npu@mail.nwpu.edu.cn). }
\thanks{ Zihao Wu, Haixing Dai, Zhengliang Liu, and Tianming Liu are with the School of Computing, The University of Georgia, Athens 30602, USA. (e-mail: \{zihao.wu1, haixing.dai, zl18864, tliu\}@uga.edu).}
\thanks{Lichao Sun is with the Department of Computer Science and Engineering, Lehigh University, PA 18015, USA. (e-mail: lis221@lehigh.edu).}
\thanks{Xiang Li is with the Department of Radiology, Massachusetts General Hospital and Harvard Medical School, Boston 02115, USA. (e-mail: XLI60@mgh.harvard.edu).}
\thanks{Dajiang Zhu is with the Department of Computer Science and Engineering, The University of Texas at Arlington, Arlington 76019, USA. (e-mail: dajiang.zhu@uta.edu).}
\thanks{Dinggang Shen is with the School of Biomedical Engineering, ShanghaiTech University, Shanghai 201210, China; Shanghai United Imaging Intelligence Co., Ltd., Shanghai 200230, China; Shanghai Clinical Research and Trial Center, Shanghai, 201210, China. (e-mail: Dinggang.Shen@gmail.com).}
\thanks{Qiushi Yang is with the Department of Electronic Engineering, City University of Hong Kong, Hong Kong 999077, China. Yixuan Yuan is with the Department of Electronic Engineering, Chinese University of Hong Kong, Hong Kong 999077, China. (e-mail: qsyang2-c@my.cityu.edu.hk; yxyuan@ee.cuhk.edu.hk).}
\thanks{Yiheng Liu and Bao Ge are with the School of Physics and Information Technology, Shaanxi Normal University, Xi’an
710119 China. (e-mail: \{liuyiheng,bob\_ge\}@snnu.edu.cn)}
\thanks{Yi Pan is with the School of Glasgow College, University of Electronic Science
and Technology of China, Chengdu 611731, China. Xi Jiang is with the School of Life Science and Technology, University of Electronic Science and Technology of China, Chengdu 611731, China.  (e-mail: dwaynepan5277@gmail.com; xijiang@uestc.edu.cn)}}

\markboth{Journal of \LaTeX\ Class Files,~Vol.~14, No.~8, August~2021}%
{Shell \MakeLowercase{\textit{et al.}}: A Sample Article Using IEEEtran.cls for IEEE Journals}

\maketitle

\begin{abstract}
Prompt engineering is a critical technique in the field of natural language processing that involves designing and optimizing the prompts used to input information into models, aiming to enhance their performance on specific tasks. With the recent advancements in large language models, prompt engineering has shown significant superiority across various domains and has become increasingly important in the healthcare domain. However, there is a lack of comprehensive reviews specifically focusing on prompt engineering in the medical field. This review will introduce the latest advances in prompt engineering in the field of natural language processing for the medical field. First, we will provide the development of prompt engineering and emphasize its significant contributions to healthcare natural language processing applications such as question-answering systems, text summarization, and machine translation. With the continuous improvement of general large language models, the importance of prompt engineering in the healthcare domain is becoming increasingly prominent. The aim of this article is to provide useful resources and bridges for healthcare natural language processing researchers to better explore the application of prompt engineering in this field. We hope that this review can provide new ideas and inspire for research and application in medical natural language processing.
\end{abstract}


\section{Introduction}
\label{sec:introduction}
Prompt engineering has emerged as a cutting-edge approach in the field of natural language processing (NLP), providing a more efficient and cost-effective means of using large language models (LLM). This innovative paradigm is rooted in the development of LLMs, which have transformed our understanding of natural language~\cite{liu2023pre,white2023prompt}.

The history of language models can be traced back to the 1950s, but it is until the introduction of the BERT and GPT models in 2018 that LLMs became the mainstream~\cite{hu2023advancing}. These models utilize complex algorithms and massive amounts of training data to understand natural language in ways that are previously unimaginable. Their ability to learn patterns and structures in language has proven invaluable in a wide range of language-related tasks~\cite{raffel2020exploring,qiu2020pre}. However, fine-tuning these pre-trained models can be a costly and time-consuming process, requiring significant amounts of annotated data and computing resources. To address this challenge, researchers have turned to prompts as a means of guiding model learning~\cite{white2023prompt,lester2021power,kopavc2002optimal}. Prompt learning is a novel paradigm in NLP that enables language models to perform few or even zero-shot learning, adapting to new scenarios with minimal labeled data~\cite{ding2022openprompt,white2023prompt,liu2023pre}. This approach is rooted in language modeling, directly modeling the probability of text. The key to prompt engineering is designing prompts for downstream tasks, which guide the pre-trained model to perform the desired task~\cite{liu2022design}.

Prompt learning can be broken down into five key steps~\cite{liu2023pre}. First, researchers must choose an appropriate pre-training model. Next, they must design prompts for downstream tasks, which can be tailored to the specific requirements of each task, and this step is called the prompt engineering process. The third step involves designing responses based on the task at hand, allowing the model to produce the desired output. The fourth step is to expand the paradigm to further improve results or adaptability methods. Finally, researchers must design training strategies that enable the model to learn efficiently and effectively. Prompt engineering is thus a critical step in the prompt-based approach, which involves designing effective prompts to guide the pre-trained language model in downstream tasks. While prompt engineering plays a crucial role in the success of the overall approach, studies have shown that the design of prompts can significantly impact the performance of the model on downstream tasks. A well-designed prompt should provide clear guidance to the model and facilitate effective task completion. Moreover, optimizing prompt parameters is also an important aspect of prompt engineering, as it can lead to improved model performance. The subsequent steps in prompt-based methods, such as designing answers, expanding the paradigm, and adjusting training strategies, require task-specific prompt design and optimization.

Overall, prompt engineering is a promising new approach in NLP that has the potential to revolutionize the field. Its focus on using prompts to guide model learning provides a more cost-effective and efficient means of training large language models, making it an increasingly important area of research in the years to come~\cite{sorensen2022information}.

\subsection{Scope and Focus of the Review}
In recent years, the healthcare industry has received increasing attention as it is closely related to each individual. However, there is still a significant gap between the shortage of medical resources and people's needs~\cite{rong2020artificial,zhou2007temporal}. NLP technology can handle massive amounts of data from various medical literature and medical records, thus helping doctors and patients better understand and manage diseases, which is of great significance in improving medical standards and providing better health security~\cite{baud1992natural}.

\begin{figure}[ht]
\begin{center}
\includegraphics[width=1.0\columnwidth]{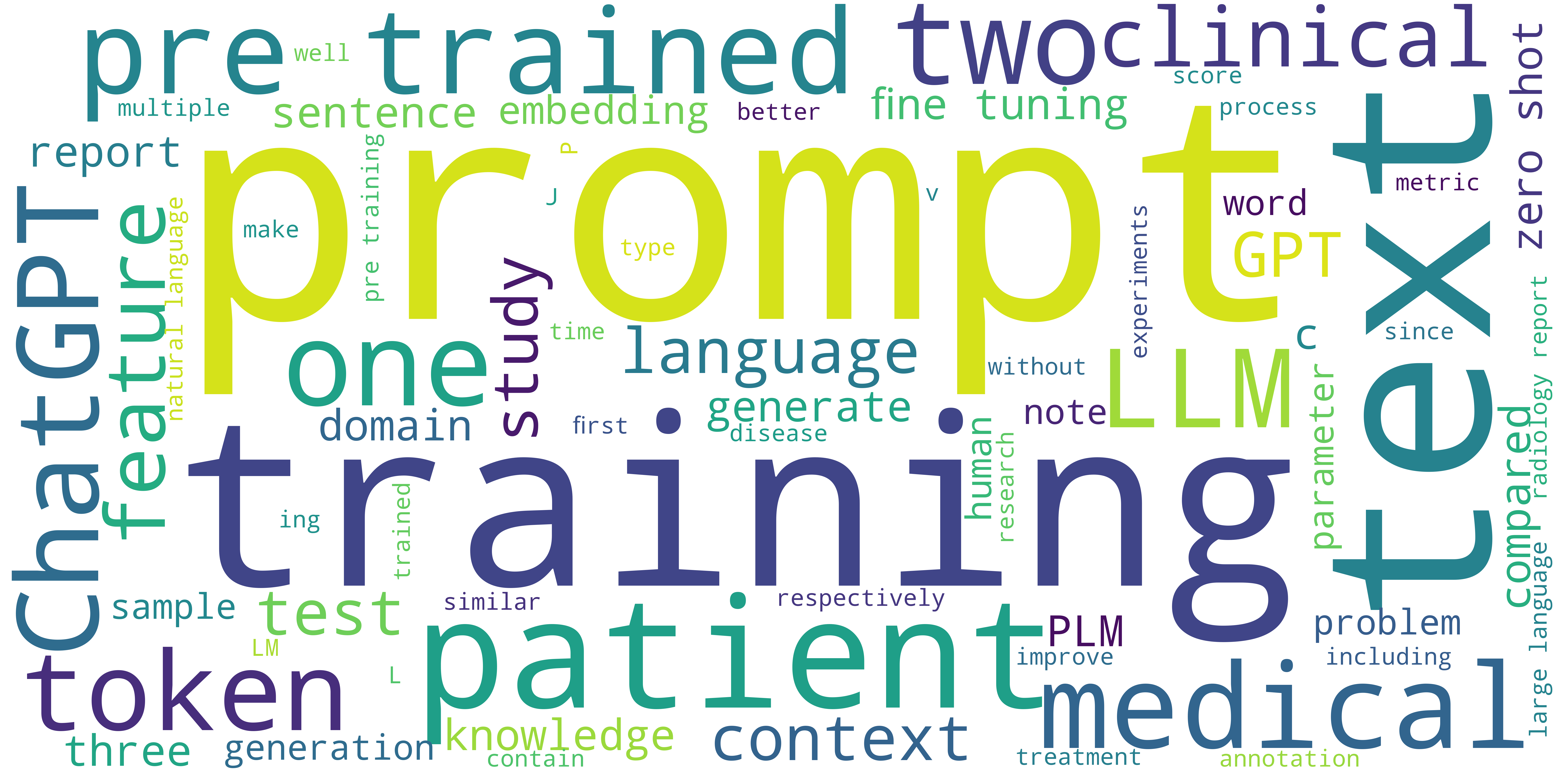}
\end{center}
\caption{A wordcloud is employed to illustrate the research hotspots, focal points, and directions in prompt engineering for NLP in the medical domain, which reflects the key vocabulary and highlights the significant themes of this literature review.
} 
\label{wordcloud}
\end{figure}

\begin{figure*}[ht]
\begin{center}
\includegraphics[width=0.9\textwidth]{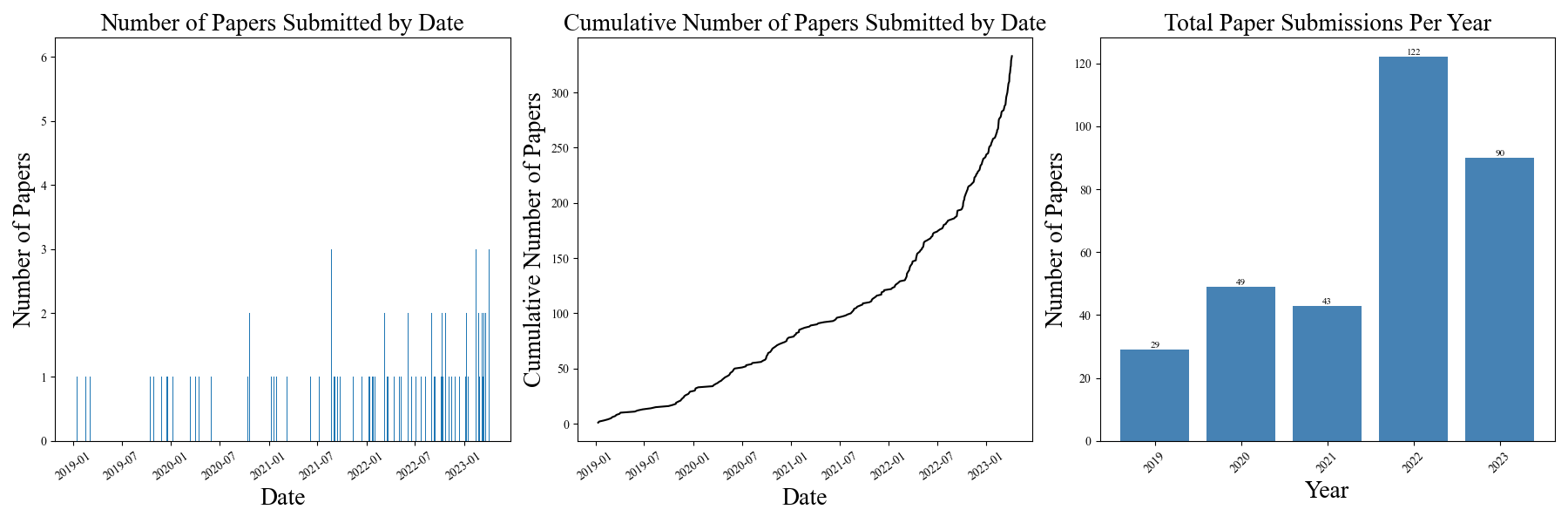}
\end{center}
\caption{The graphical representation is utilized to depict the number of research papers on prompt engineering for NLP in the medical domain, published from 2019 to April 6, 2023, revealing the trend and growth of this field over time. The graph showcases three different plots: daily submitted count, cumulative daily submitted count, and annual submitted count.}
\label{statistics}
\end{figure*}

\begin{figure*}[!h]
\begin{center}
\includegraphics[width=0.9\textwidth]{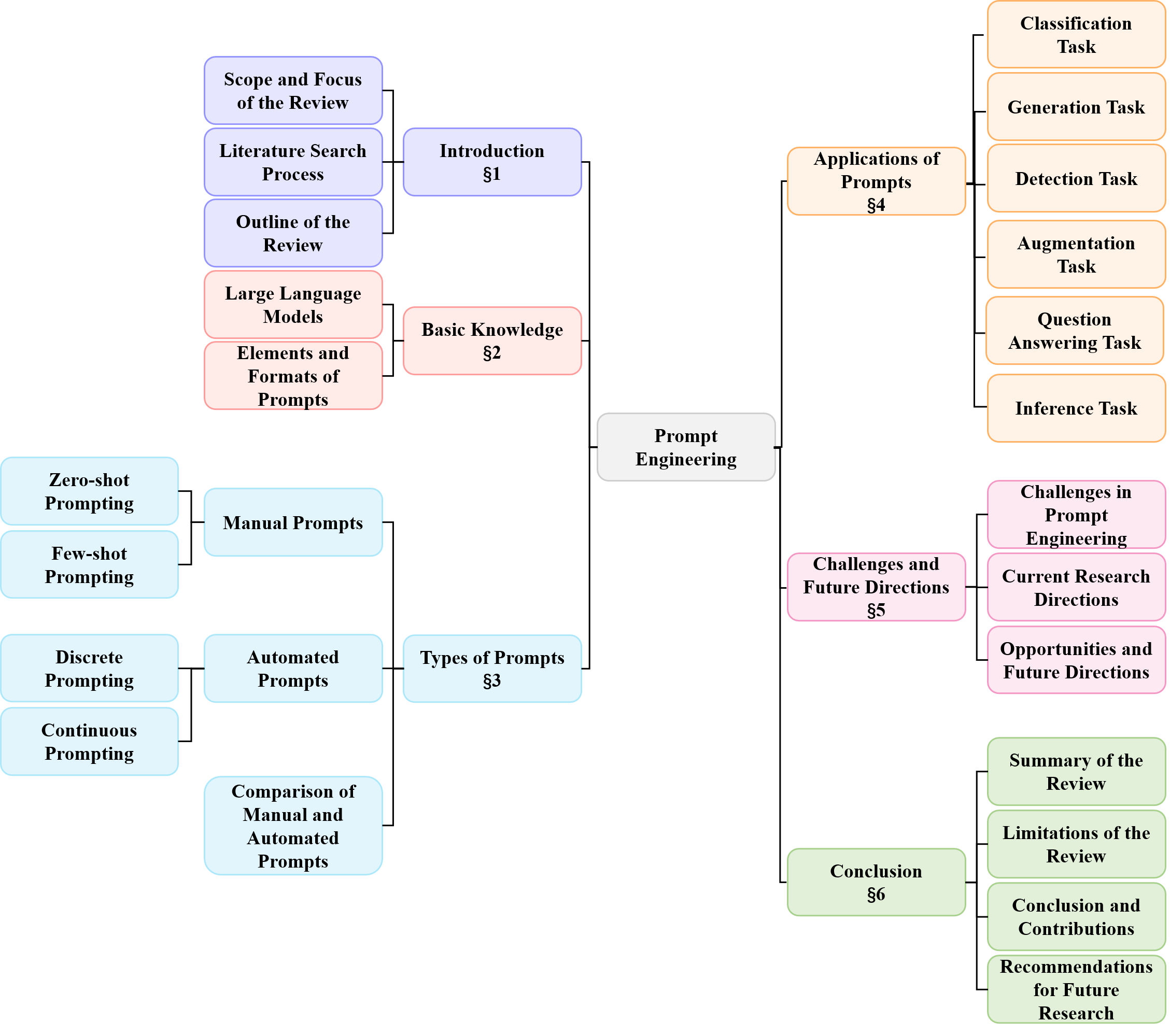}
\end{center}
\caption{The overview of this literature review provides a comprehensive and systematic summary of the current state of research on prompt engineering for NLP in the medical domain.
} 
\label{outline}
\end{figure*}

However, traditional machine learning and deep learning methods cannot solve NLP tasks in the medical field very well~\cite{hu2023advancing}. The complex and diverse professional terminology in the medical field, the wide range of professional knowledge, and ethical issues related to patient privacy have all become difficult problems in the development of the healthcare industry~\cite{liu2023deid}. The emergence of LLMs and prompt-based methods provides a new solution for NLP tasks in the medical field. As depicted in Figure \ref{wordcloud}, the wordcloud illustrates the research hotspots and focal points in NLP for the medical domain, highlighting the significant themes of this literature review in this field. Through the powerful contextual learning ability of LLMs, useful information can be effectively obtained from a large number of medical literature and cases. By designing specific prompts, domain-specific knowledge and task-specific information can be introduced into the pre-trained model, resulting in better performance and gaining more extensive attention~\cite{dai2023chataug}.

In the research of prompt engineering, designing appropriate prompts is an important issue. Currently, researchers are exploring various types of prompts, including manual prompts~\cite{petroni2019language,wei2023zero,liu2023deid,dai2023chataug,lyu2023translating,lamichhane2023evaluation} and automated prompts~\cite{wallace2019universal,shin2020autoprompt,li2021prefix,lester2021power}, as well as various methods of prompt construction to help solve tasks in the medical field.

\subsection{Literature Search Process}
The importance of NLP tasks in the medical field cannot be overstated. With the increasing volume of medical data, there is a growing need for effective NLP techniques to improve the quality and efficiency of medical services. In this regard, prompt engineering has emerged as a promising approach to guiding model generation by providing targeted prompt information.

To investigate the application of prompt engineering in the medical field, we conduct a comprehensive review of literature from 2019 to 2023 by searching for keywords ``prompt" and ``medical NLP" on arxiv. As shown in Figure \ref{statistics}, a total of 333 papers related to the prompt are identified, and ChatGPT is used to screen the abstracts for their relevance to the medical field, resulting in the selection of 140 relevant papers for further review. The dynamic graph in the following link\footnote{https://braininspiredai.github.io/prompt\_trend\_in\_medical} shows the daily publication rate of related papers.

These papers mainly focus on the design and application of prompt engineering in NLP tasks in the medical field, with the goal of providing suitable prompt design methods for different medical tasks. Specifically, the studies explore how to choose and design prompt elements, how to use prompts to guide a model generation of text that meets medical requirements, and how to evaluate the impact of different prompt designs on model performance.

\subsection{Outline of the Review}
This review article provides an in-depth overview of Prompt Engineering, a rapidly growing field of research focusing on enhancing medical applications through the development of effective prompts. As illustrated in Figure \ref{outline}, we present the overview of this review article with each section organized as follows:

The introduction of the article discusses the background and significance of Prompt Engineering and outlines the scope and focus of the review. It describes the literature search process followed and provides an outline of the article.

Section II presents the basics of Prompt Engineering, including common LLMs, and the elements and components of prompts.

Section III discusses the different types of prompts available in the literature, with a focus on manual prompts such as zero-shot and few-shot prompting. The section uses medical literature to highlight specific details. Automated prompts, including discrete and continuous prompting, are also discussed, and the section compares manual and automated prompting techniques.

In Section IV, the various applications of prompts in medical fields are covered in detail. These applications include classification, generation, detection, augmentation,  question answering, and inference tasks, and the section provides the latest medical examples to illustrate the practical use of prompts.

Section V outlines the challenges and future directions of prompt engineering in medical applications. The section highlights the current research directions in the field and provides insight into the opportunities for future research.

Finally, Section VI concludes the review by summarizing the key findings and contributions of the article. The section also discusses the limitations of the review and provides recommendations for future research in the field of prompt engineering in medical applications.

Overall, the review provides insights into the latest developments in prompt engineering for medical NLP tasks and highlights the importance of prompt engineering in improving the accuracy and effectiveness of medical services. It also underscores the need for continued research to explore the potential of prompt engineering in the medical field and to identify effective strategies for its implementation. Ultimately, prompt engineering has the potential to transform the way medical services are delivered and to enhance the quality of care for patients.

\section{Basic Knowledge}
In prompt engineering, LLMs are of great importance as they enable the design of appropriate prompts for specific tasks and can generate high-quality results even with limited training data~\cite{petroni2019language}. This section will delve into the latest advances in large models and provide a framework for prompt design.

\subsection{Large Language Models}
In this section, we have explored the prevalent LLMs in prompt engineering within the medical field, which are trained on large amounts of text data and are capable of generating high-quality, contextually relevant, and coherent text in various NLP tasks. Among them, prompt-based techniques provide researchers with rich natural language patterns and structures and have become an important tool in prompt engineering.

Prompt engineering is achieved through the design of prompts, which enables the model to generate diversified text based on different contextual environments and can be optimized and customized for different tasks and application scenarios~\cite{zhou2022learning}. In the medical field, the application of these models has provided significant help and inspiration for medical research and clinical practice.

BERT (Bidirectional Encoder Representations from Transformers)~\cite{kenton2019bert} is a pre-trained model developed by the Google team in 2018 that has bidirectional encoder representations using the transformer architecture. Its purpose is to pre-train bidirectional representations by jointly adjusting the context in all layers, and by overcoming the limitations of previous unidirectional language models through a masked language model. In addition, BERT introduces the next sentence prediction task, which can be used in conjunction with a masked language model to pre-train text pair representations~\cite{he2022masked}. The training objective of BERT is highly meaningful, as it demonstrates the importance of bidirectional pre-training for language representations. Compared to previous unidirectional language models, BERT's pre-training can better capture semantic information in the context~\cite{kenton2019bert}.

BERT is the first fine-tuning based representation model, which has surpassed many task-specific architectures in 11 NLP tasks, setting new performance records~\cite{kenton2019bert}. This achievement demonstrates the enormous potential of pre-trained models in NLP tasks, and provides important references for the development of subsequent NLP models.

The Text-to-Text Transfer Transformer (T5)~\cite{raffel2020exploring} is an LLM developed by Google in 2019. The fundamental idea behind T5 is to treat every NLP task as a ``text-to-text" problem. The training objective of T5 is to develop it into a universal knowledge representation model, enabling the model to better understand and process text. T5 has achieved state-of-the-art results in multiple NLP tasks, such as text classification, machine translation, text summarization, and question answering, among others. In these tasks, the text is taken as input, and new text is generated as output to solve specific problems. The advantage of this approach is that by treating all NLP tasks as ``text-to-text" problems, the T5 model can learn how to map input text to output text, thereby enhancing its understanding and processing of text.

ChatGPT is a series of generative pre-trained transformer models developed by the OpenAI team. GPT-1~\cite{liugenerating}, which is released in June 2018, is trained on a large-scale text corpus with the aim of generating coherent and natural text. Subsequently, GPT-2~\cite{radford2019language} is released in February 2019, which is built on the foundation of GPT-1 with a larger model size and more training data, achieving outstanding performance on a variety of NLP tasks. In June 2020, GPT-3~\cite{brown2020language} is introduced, utilizing an even larger model size and more extensive training data than GPT-2, resulting in further improvements in model performance. GPT-3.5 then further optimizes the efficiency and performance of GPT-3.

To enhance the performance of the model, the OpenAI team uses supervised and reinforcement learning to fine-tune GPT-3.5, with human intervention playing a key role in enhancing machine learning capabilities~\cite{najar2021reinforcement,knox2009interactively}. The optimization of training methods and increased training data has led to GPT-4 being able to generate text that is similar to that produced by humans, representing a significant breakthrough in the field of artificial intelligence.

In the future, LLMs will continue to evolve and improve, and we can expect their applications in the medical field to bring more value and contributions to human health.

\begin{figure}[ht]
\begin{center}
\includegraphics[width=1.0\columnwidth]{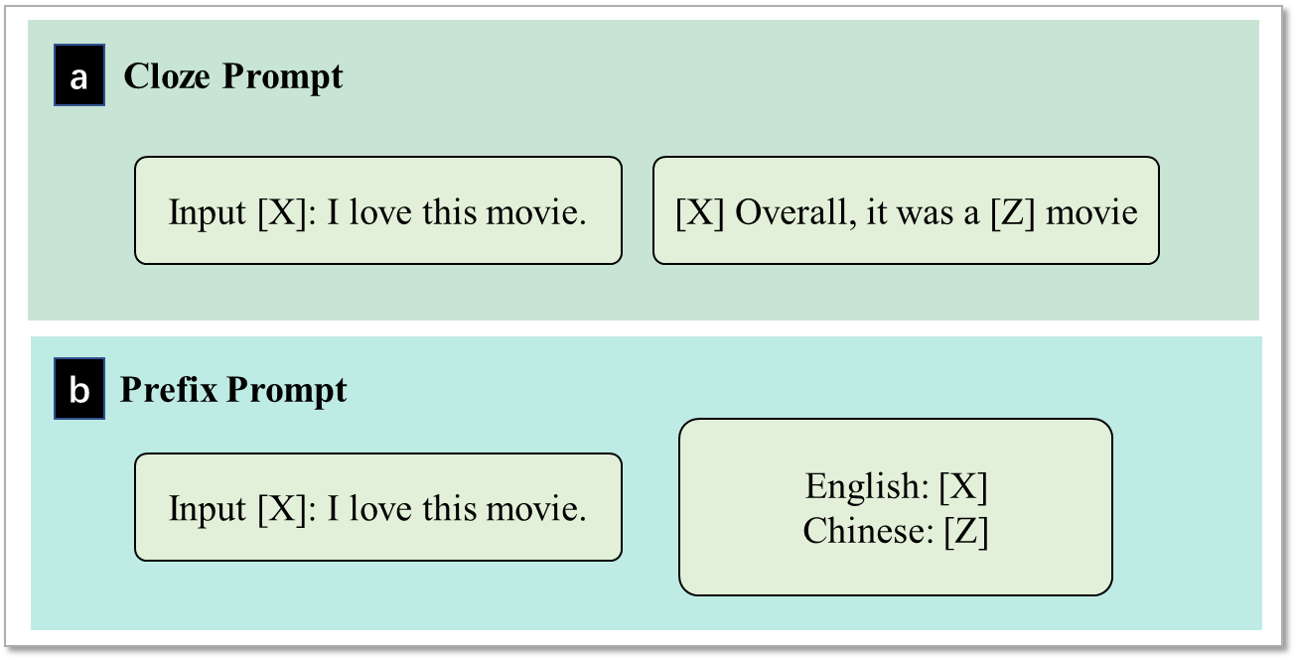}
\end{center}
\caption{This study distinguishes between two types of prompts, namely cloze prompts (illustrated in a) and prefix prompts (illustrated in b), where slot [x] represents the input text and slot [z] represents the answer text.
} 
\label{cloze and prefix}
\end{figure}

\subsection{Elements and Formats of Prompts}
Prompts are usually classified into two forms: cloze prompts~\cite{petroni2019language,cui2021template} and prefix prompts~\cite{li2021prefix,lester2021power}. Cloze prompts refer to a slot that needs to be filled in the middle of the template text. For example, in sentiment analysis, when X = ``I love this movie," the template may take the form of `` [X]. Overall, it was a [Z] movie." Then, the new sentence X' becomes ``I love this movie. Overall it was a [Z] movie." This is an example of cloze prompts as shown in Figure \ref{cloze and prefix}(a). Prefix prompts, on the other hand, refer to the input text being entirely situated before the generated answer text Z. As shown in Figure \ref{cloze and prefix}(b), ``English: [X] Chinese:[Z]".

Prompt engineering refers to the process of creating a prompt function that guides a model to perform effectively in downstream tasks. This process typically involves two steps. First, a template is applied, which is a text string that contains two slots: an input slot [X] for the input text x and an answer slot [Z] for the intermediately generated answer text z that will later be mapped into the final output y. Then, the input slot [X] is filled with the input text x~\cite{liu2023pre}.

It is important to note that in prompt engineering, prompts are not only in the form of natural language but can also be generated continuous vector embeddings~\cite{li2021prefix}. This part will be discussed in detail in section three. By utilizing prompt engineering techniques, model performance and efficiency can be significantly improved, providing better support and assistance for downstream tasks.

\section{Types of Prompts}
\begin{figure}[ht]
\begin{center}
\includegraphics[width=1.0\columnwidth]{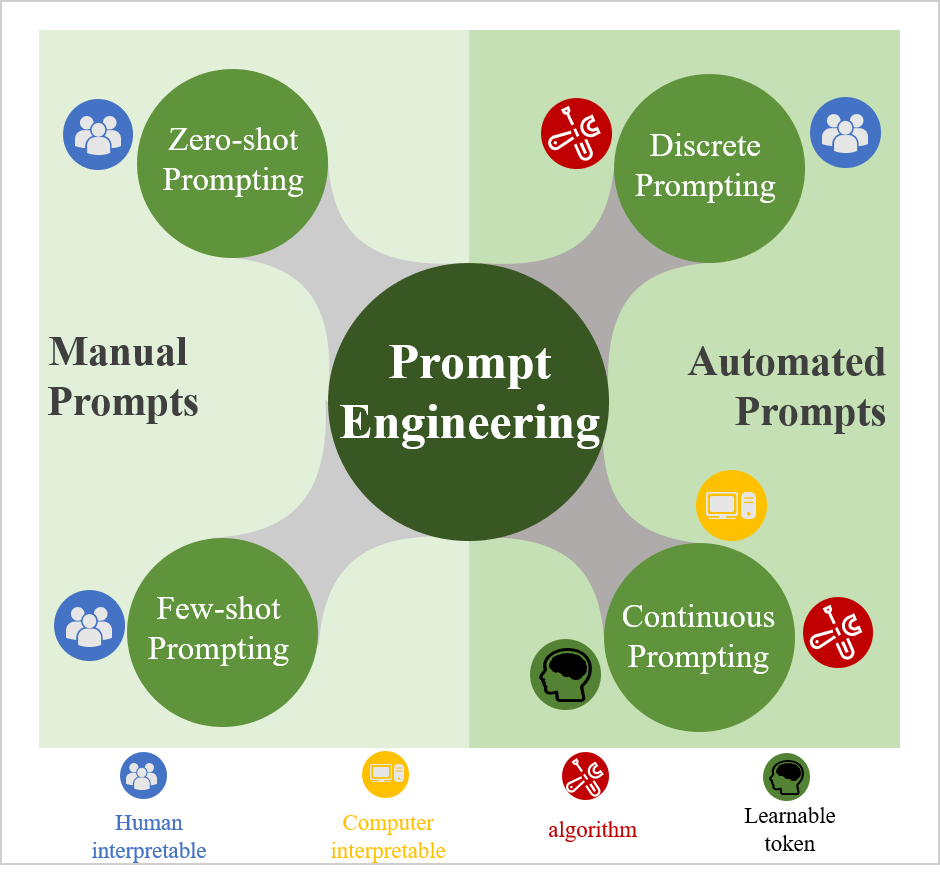}
\end{center}
\caption{Prompt engineering can be broadly categorized into two types: manual prompts and automated prompts. Manual prompts encompass zero-shot prompting and few-shot prompting, both of which rely on human expertise for their manual configuration. On the other hand, automated prompts consist of discrete prompting and continuous prompting, which involve the design of a series of automatic algorithms. Discrete prompts are typically human interpretable, whereas continuous prompting usually employs learning tokens that are interpretable by computers.
} 
\label{prompt type}
\end{figure}

In the realm of scientific research, prompts play a crucial role in guiding large models to produce accurate outputs. As shown in Figure \ref{prompt type} Prompts are broadly categorized into two types: manual prompts and automated prompts~\cite{liu2023pre}. Manual prompts are meticulously crafted by human experts to provide explicit instructions to the model on what type of data to focus on and how to approach the task at hand. These prompts are particularly effective when the input data is well-defined and the output must conform to a specific structure or format.

However, manual prompts have certain limitations that cannot be overlooked. Creating effective manual prompts requires significant expertise and time, and even minor changes to the prompts can have a great impact on model predictions~\cite{petroni2019language,wei2023zero,liu2023deid,dai2023chataug,lyu2023translating,lamichhane2023evaluation}. This is especially true for complex tasks where providing clear and effective prompts is a challenging endeavor. To address these limitations, researchers have developed various automated methods to design prompts.

Automated prompts have gained popularity due to their efficiency and adaptability. These prompts are generated using various algorithms and techniques, eliminating the need for human intervention~\cite{wallace2019universal,shin2020autoprompt,li2021prefix,lester2021power}. Discrete prompts and continuous prompts are two common types of automated prompts. Discrete prompts rely on predefined categories to generate responses, while continuous prompts consider the current conversation context to produce accurate outputs~\cite{liu2023pre,sivarajkumar2022healthprompt,milecki2023medimp,feng2022beyond,yang2023bliam}. Additionally, there are static prompts and dynamic prompts that vary in their consideration of historical context~\cite{liang2022modular,gao2023prefixmol}.

\subsection{Manual Prompts}
Manual prompts can be used to guide models towards producing the desired results for specific tasks. These prompts are often created based on human expertise, and are commonly used in LLMs to improve their performance on downstream tasks. For instance, the LAMA dataset introduced hand-crafted cloze prompts to explore the knowledge of large models~\cite{petroni2019language}. Similarly, prefix prompts can be manually created to facilitate common sense reasoning tasks. In Chat-oriented models like ChatGPT, zero-shot prompting is used to guide the models towards producing ideal results for downstream tasks. Additionally, it is common to include prompt examples that match the level of task complexity, to further guide models in producing accurate outputs. In this section, we mainly discuss the few-shot prompting that is manually created through instruction with examples selected by human experience, rather than being automatically selected. As AI technology rapidly advances, automated few-shot prompting techniques that select examples automatically are also emerging, which will be discussed in section 3.2.

\subsubsection{Zero-shot Prompting}
The superior performance of LLMs in a wide range of NLP tasks has garnered significant attention due to their remarkable ability to learn from context, making them an attractive tool for various research problems. This emerging capability is often enhanced through prompt engineering to improve the performance of LLMs in downstream tasks. Consequently, researchers have proposed various methods for designing prompts to guide the models' applications in different research questions~\cite{abaho2022position}.

Recently, the development of LLMs such as GPT-3 and ChatGPT has made the use of prompt-based methods increasingly popular for various downstream tasks. Zero-shot prompting, in particular, has shown great promise, where providing a well-designed prompt alone without corresponding examples can lead to outstanding results. Some studies have suggested that leveraging the powerful performance of GPT can help us complete downstream tasks such as extraction and recognition, and even outperform some full-shot models on certain datasets~\cite{wei2023zero}. In the medical field, the impressive achievements of using LLMs to exploit their strong contextual capabilities have also been observed. For instance, DeID-GPT proposes the use of high-quality prompts to preserve privacy and summarize key information in medical data on ChatGPT and GPT-4, achieving better results compared to several baseline methods. The study provides a foundation for future research on applying medical data to LLMs~\cite{liu2023deid}. ChatAug proposes a prompt-based medical data augmentation method on ChatGPT, which outperforms other popular methods. The study highlights the importance of domain knowledge in generating correct augmented data and suggests a fine-tuning approach for future research~\cite{dai2023chataug}. Recent work has investigated the feasibility of using manual prompts to guide downstream tasks on ChatGPT and has shown that ChatGPT's performance on translation tasks can be significantly improved with clear and accurate prompts~\cite{lyu2023translating}. Similarly, research on ChatGPT's zero-shot prompting has also yielded similar conclusions~\cite{lamichhane2023evaluation}. The study presents HealthPrompt, a prompt-based clinical NLP framework that explores different types of prompt templates, including prefix prompts and cloze prompts, to improve zero-shot learning performance on clinical text classification tasks without requiring additional training data~\cite{sivarajkumar2022healthprompt}. These findings highlight the potential of ChatGPT to enhance models' performances on NLP tasks with effective prompt design, providing valuable insights for future research in this area.

Overall, the success of prompt-based methods has shown their great potential for further advancements, and their significance is likely to continue to grow in the future.

\subsubsection{Few-shot Prompting}

While zero-shot prompting has shown impressive performance in many tasks, it still has some limitations. First, it heavily relies on the performance of pre-trained models. Secondly, due to the flexibility of zero-shot prompting, its outputs may not always be accurate. In such cases, providing a small number of prompt examples as guidance for the model to achieve better performance has become a useful approach. Few-shot prompting can serve as clear and explicit prompt inputs to guide the model towards desired outputs. For instance, research has been conducted to measure the baseline performance of GPT-4 on medical multiple-choice questions (MCQs) using only a few prompt samples, without the need for complex methods such as chain-of-thoug~\cite{lai2023chatgpt}.

These prompt methods rely on the contextual emergence ability of LLMs, which has demonstrated impressive performance on ChatGPT/GPT-4~\cite{holmes2023evaluating}. The manual prompt-based approach has shown typical few-shot prompting capabilities in the medical text translation, text augmentation, generation, summarization, and other tasks, with significant improvements over baseline models on public datasets~\cite{lamichhane2023evaluation,yuan2023llm}.

\subsection{Automated Prompts}
The effectiveness of human-designed prompts depends on the prompt designer's expertise in selecting relevant information and constructing prompts in a way that is understandable to the model. Human prompts are particularly effective for tasks with well-defined input data and structured output requirements. However, designing effective human prompts requires a significant amount of time and domain-specific knowledge, making their usage in more complex tasks challenging. Therefore, researchers are exploring automated prompt design methods to address these challenges and improve the efficiency and adaptability of prompt-based approaches.

\subsubsection{Discrete Prompting}
Discrete prompts refer to automatically searching for templates in a discrete space, typically corresponding to natural language phrases, to provide the model with guidance for generating the desired output. This approach can help us design prompts more easily and improve the efficiency and adaptability of the model~\cite{liu2023pre}. Common discrete prompt construction methods include prompt mining, prompt paraphrasing, prompt generation, and prompt scoring~\cite{zhou2022large}.

Prompt mining is a method of automatically discovering templates or prompts by scraping a large text corpus for frequent middle words or dependency paths between a set of input-output pairs. These prompts can be used for various natural language processing tasks such as language modeling, text classification, and question answering. The research proposes a method for prompt-based learning that automatically identifies and adds prompt phrases to a prompt template for fine-grained detection of Alzheimer's disease. The method enhances detection accuracy through the use of additional prompts~\cite{wang2022exploiting}.

Prompt paraphrasing involves generating a set of candidate prompts from an existing seed prompt using methods such as round-trip translation or phrase replacement, and selecting the one that achieves the highest training accuracy for a target task. This approach can be optimized using a neural prompt rewriter and can be done on an individual input basis. For instance, the article proposes a simple yet effective prompt method based on paraphrasing to assist pre-trained models in learning rare biomedical vocabulary, resulting in improved performance in biomedical applications~\cite{wang2022prompt,romanov2018lessons,gao2021making}. The STREAM framework leverages prompt-based language models to generate task-specific logical rules for named entity tagging, reducing the need for human labor. By utilizing existing prompt templates and continuously teaching the language model, STREAM achieves higher accuracy and efficiency in the tagging process~\cite{chen2022distilling}.


Prompt generation is the task of generating prompts using natural language generation models. The study proposes MEDIMP, a framework that generates medical prompts using automatic textual data augmentations from LLMs, and demonstrates its superiority in producing high-quality prompts for medical applications~\cite{milecki2023medimp}. The paper proposes a method that utilizes prompts filled with bounding box annotations to generate descriptions containing extensive hints and context for instance recognition and localization. The knowledge from language is then distilled into the detection model via maximizing cross-modal mutual information in both image and object levels~\cite{feng2022beyond}. This approach involves training prompts from existing data and using these prompts to generate new data. This newly generated data can then be used to define additional prompts, ultimately providing downstream tasks with more high-quality feature representations~\cite{yang2023bliam}. The study introduces a prompt construction module that leverages medical language templates to enable pre-trained language models to extract contextual information from tabular electronic health records and generate more comprehensive cell embeddings, which can be used by a pre-trained sentence encoder to generate sentence embeddings as cell representations~\cite{ruan2023medical}.

Prompt scoring involves using language models to score filled prompts for a given task and selecting the one with the highest probability. This can result in a custom template for each individual input, as in the case of knowledge-based completion. For instance, ImpressionGPT has proposed dynamic prompts, which select the top-k examples with the highest similarity for each input. Based on the probability and threshold of the k-selected examples, the prompt for each input is determined. This approach, which utilizes few-shot learning, has shown better performance than some full-shot learning methods~\cite{ma2023impressiongpt}.

The utilization of various prompt types can greatly benefit language models in their performance of a diverse range of tasks, such as natural language understanding, text generation, machine translation, and question-answering. Each type of prompts, whether generated through prompt mining, prompt paraphrasing, prompt generation, or prompt scoring can be used independently or in combination to optimize the performance of the language model for a particular task. This underscores the significance of prompt-based approaches in advancing the capabilities of NLP systems.

\subsubsection{Continuous Prompting}
In recent years, the development of LLMs has made prompt-based methods increasingly popular in various downstream tasks. In this context, continuous prompting (e.g., soft prompting) has emerged as a new type of prompts and has gained widespread attention from researchers. Unlike discrete prompts, continuous prompts can be operated in the embedding space and are no longer limited to text-readable types. Additionally, continuous prompts can optimize parameters through tuning on training data for downstream tasks, which relaxes the constraints on prompts and improves the efficiency of LLMs in task execution. Meanwhile, continuous prompting has been shown to be an effective alternative to standard model fine-tuning in cases where the data is highly imbalanced~\cite{elfrink2023soft}. Discrete prompts are composed of discrete words or phrases, typically in human-interpretable natural language, to guide the model in performing downstream tasks. On the other hand, continuous prompts directly prompt the model in the embedding space, without constraints on natural language, and allow the prompts to have their own parameters that can be fine-tuned using training data. 

In the medical field, the advantages of continuous prompting have been further demonstrated. Liang et al. use continuous prompts to introduce prompting into the model architecture as the only trainable parameter to guide model output, proposing PromptFuse and BlindPrompt as methods for aligning different modalities in a modular and parameter-efficient manner. These methods require only a few trainable parameters and perform comparably to several mutil-modal fusion methods in low-resource scenarios. The high modularity property of prompting allows for the flexible addition of modalities at low cost, as it avoids the need to fine-tune large pre-trained models~\cite{liang2022modular}. PrefixMol is a generative model that utilizes a learnable prompt token as prefix embedding to guide the model to generate an output that meets the desired criteria. This prompt token encodes a set of learnable features and contextual information about the targeted pocket's circumstances and various properties. This approach offers significant advantages over traditional generative models as it allows for more precise and efficient control over the output~\cite{gao2023prefixmol}. Wang et al. propose an adaptive PromptNet for glioma grading that utilizes only non-enhanced MRI data and receives constraints from features of contrast-enhanced MR data during training through a designed prompt loss. In the paper, enhanced features are used as prompts to guide the feature extraction of PromptNet, with an adaptive strategy designed to dynamically weight the prompt loss in a sample-based manner, thus achieving competitive glioma grading performance on NE-MRI data~\cite{wang2022adaptive}. CPP utilizes task-specific prompts optimized with a contrastive prototypical loss to avoid semantic drift and prototype interference, using prompts as learnable tokens to train task-specific information for each task. It also introduces a multi-centroid prototype strategy to improve prototype representativeness. CPP outperforms existing methods under a light memory budget and improves class separation~\cite{li2023steering}.

Overall, continuous prompting is a powerful tool for improving the performance of automated systems. It helps these systems to learn and adapt more quickly, and to perform more accurately in a wide range of applications. This technique can be used in combination with other types of prompts, such as manual prompts and discrete prompts, to create a more comprehensive and effective learning environment, which is also used in multi-modal tasks such as image and speech recognition to improve the accuracy and speed of the systems. Additionally, continuous prompts can be tailored to the specific needs of the task, which makes it highly adaptable and versatile.

\subsection{Comparison of Manual and Automated Prompts}

Manual prompts refer to human-crafted prompts, while automated prompts are generated by algorithms or automated methods. The primary difference between manual and automated prompts is the level of human involvement in the prompt construction process. Manual prompts are carefully designed by human experts, while automated prompts are generated using various search algorithms or other automated methods. The advantage of manual prompts is that they are typically well-designed and can capture important aspects of the downstream task. However, the manual prompt design process can be time-consuming and costly. On the other hand, automated prompts are generated more quickly and can potentially capture more task-specific information than manually crafted prompts. However, the effectiveness of automated prompts heavily depends on the quality of the search algorithms and the representation power of the language models.

In summary, both manual prompts and automated prompts have their advantages and disadvantages. Manual prompts provide greater control over the output, while automated prompts are more efficient and adaptable. Ultimately, the choice of the prompt will depend on the specific task and available resources.

\section{Applications of Prompts}
\label{sec:guidelines}

Prompt engineering enables language models to achieve a wide range of AI tasks in the medical field with high performance. With customized medical prompts, models can accomplish complex healthcare problems that are previously difficult without huge medical datasets and resources. 
In the medical field, the applications of prompt engineering include:

\begin{itemize}
    \item Classification \cite{zhang2023gpt4mia, sivarajkumar2022healthprompt, lamichhane2023evaluation, akrout2023diffusion, zhu2023segprompt, elfrink2023soft, wang2022exploiting, chen2023towards, lin2023exploring}: Medical prompts elicit predicted diagnoses, categories or tags from models. For example, ``The patient's CT scan results show: [tumor type]" can steer a model to generate ``thymoma", etc. Classification prompts provide context for model predictions in the medical domain. They are useful for condition classification, diagnosis classification, etc.
    
    \item Generation \cite{chambon2022adapting, lu2023napss, lyu2023translating, wang2023chatcad, lehman2022learning, abaho2022position, weber2023cascaded, lee2023clinical}: Open-ended medical prompts give models freedom to generate creative samples like case reports, draft research papers, and so on, from their knowledge. For instance, ``Here is a possible case report:" can lead to an original case report. Generation prompts allow unconstrained production of new medical data samples. They are applicable for creative medical tasks like clinical case modeling and medical article writing.  
    
    \item Detection \cite{qin2022medical, liu2023deid, ye2023uniseg}: Medical prompts frame object or anomaly detection contexts and ask models to report entities detected. For example, ``Detect any abnormal findings you observe in the attached CT images". Detection prompts imply what medical models need to identify. They are relevant for locating and classifying diseases, anatomical anomalies or other irregularities in medical images, videos, and documents. 

    \item Augmentation \cite{dai2023chataug, milecki2023medimp, yuan2023llm, li2023semantic}: Medical prompts provide additional context to augment data for training language models. For example, we can input samples of all classes into language models like ChatGPT and prompt the model to generate samples that preserve semantic consistency with existing labelled data. The generated data, together with original samples, are used to train classifier models. Augmentation prompts add details to expand medical datasets. They are useful for boosting model performance on certain medical tasks by synthesizing more data context. 

    \item Question Answering \cite{singhal2022large, van2023open, li2022towards, kasai2023evaluating, holmes2023evaluating, jang2023exploring, zuccon2023dr, nori2023capabilities}: Medical prompts pose questions for models to comprehend and accurately respond to based on medical knowledge. For example, ``What year was radiocontrast invented?" prompts the model to search its medical knowledge base and answer ``1923". Question answering prompts assess if medical models understand questions and can generate correct responses by finding relevant medical information or facts.
    
    \item Inference \cite{wu2023exploring, long2023can, wang2022prompt}: Medical prompts present scenarios and have models explain their clinical reasoning or diagnostic process. For example, ``Here are two medical variables: A = [some values], B = [some values]. Explain how A affects B and why." The model must logically relate A and B based on medical knowledge. Reasoning prompts require coherent medical explanations and inference generation. They aim to test clinical logic, causality reasoning, and argumentation abilities of medical models.


\end{itemize}

In summary, prompt engineering provides a framework of adapting and improving language models for diverse medical applications. Prompt design is the key to unlocking model capabilities for different medical problems. 

\subsection{Classification Task}
Prompt engineering techniques show great promise for medical classification tasks when annotated data is scarce. Some studies apply prompt-based learning with pre-trained language models for clinical and mental health classification. As shown in Table \ref{classification}.

\begin{table*}[!h]
\caption{Applications of prompt engineering in classification tasks}
\centering
\begin{tabular}{p{2.5cm}p{3.5cm}p{1.5cm}p{4cm}p{4cm}}
\toprule
\textbf{Reference} & \textbf{Task}                                                                                        & \textbf{Prompt type} & \textbf{Dataset}                                                                                                                                                                                                                                          & \textbf{Highlight}                                                                        \\ \midrule
Yang et al.~\cite{yang2023multi}                   & Multi-label classification                                                                          & Discrete      & US Veterans Health Administration Corporate Data Warehouse                                                                                                                                                                                                & autoregressive model                                                                      \\
Wang et al.~\cite{wang2022adaptive}                   & Auxiliary glioma diganosis                                                                           & Continuous    & BraTS                                                                                                                                                                                               & PromptNet                                                                                 \\
Kolluru et al.~\cite{kolluru2022covid }                  & Text classification and generation                                                                  & Discrete      & PRONCI                                                                                                                                                                                                                                                    & MTGEN and UNIGEN model                                                                    \\
Niu et al.~\cite{niu2023ct}                 & Lung cancer diagnosis                                       & Continuous    & NLST, LUNA16, LUNG-PET-CT-Dx2                                                                                                                                                                                                                               & CLIP and BioGPT                                                                           \\
Sivarajkumar et al.~\cite{sivarajkumar2022healthprompt}                   & Clinical text classification                                                                & Discrete      & MIMIC-III                                                                                                                                                                                                                                                 & Defining a prompt template                                                                \\
Yang et al.~\cite{yang2023dual}                   & Few-shot dialogue state tracking                                                                     & Discrete      & MultiWOZ 2.0 and 2.1                                                                                                                                                                                                                                      & SOLOIST as base model                                                                     \\
Min et al.~\cite{min2021noisy}                   & Few-shot text classification                                                                         & Manual        & SST-2, SST-5, MR, CR, Amazon, Yelp, TREC, AGNews, Yahoo, DBPedia and Subj                                                                                                                                                                                 &                                                                                           \\
Akrout~\cite{akrout2023diffusion}                   & Data augmentation, Skin condition classification                                                     & Manual        & LAION datasets                                                                                                                                                                                                                                            & A Clip-based text encoder, U-net based latent space generator and VAE based image decoder \\
Zhu et al.~\cite{zhu2023segprompt}                   & Kidney stone classification                                                                          &                      & Collect 1496 kidney stone images from 5 different videos.                                                                                                                                                                                                 &                                                                                           \\
Lamichhane~\cite{lamichhane2023evaluation}                   & Mental health classification                                                                          & Manual         & Stress Detection Dataset, Depression Detection Dataset, Suicidality Detection Dataset                                                                                                                                                                   & ChatGPT (GPT-3.5)                                                                         \\
Deyoung et al.~\cite{deyoung2020evidence}                   & Disease inference                                                      & Discrete       & The Evidence Inference dataset                                                                                                                                                                                                                           & fine-tuned BERT                                                                           \\
Elfrink~\cite{elfrink2023soft}                   & Early prediction of lung cancer                                                                      & Continuous     & Data from patients of General Practitioners. & Transformer-based pretrained language models                                              \\
Wang et al.~\cite{wang2022exploiting}                   & Alzheimer's disease diagnosis                                                                        & Discrete       & ADReSS20 Challenge dataset                                                                                                                                                                                                                                & BERT and RoBERTa                                                                          \\
Lin et al.~\cite{lin2023exploring}                   & Multiple instance learning                                        & Continuous     & Camelyon16. TCGA-NSCLC                                                                                                                                                                                                                                  &                                                                                           \\
Chen et al.~\cite{chen2023towards}                   & Medical vision-and-language pre-training & Continuous     & ROCO, MedICaT, MIMIC-CXR                                                                                                                                                                                                                                    &                                                                                           \\
Taylor et al.~\cite{taylor2022clinical}                   & Clinically meaningful decision                                                               &                      & MIMIC-III                                                                                                                                                                                                                                                 &                                                                                           \\
Yao et al.~\cite{yao2022extracting}                   & Distinguishing Disease Symptoms                                                                                                     & Manual         & BioLAMA                                                                                                                                                                                                                                                   &                                                                                           \\
Keicher et al.~\cite{keicher2022few}                   & Diagnosis of disease;                                       & Continuous     & MIMIC-CXR-JPG v2.0.0                                                                                                                                                                                                                                      & CLIP                                                                                      \\
Diao et al.~\cite{diao2022molcpt}                    & Molecular property prediction                                                                        & Continuous     & BBBP, BACE, ClinTox, Tox21, SIDER, HIV, MUV, and ToxCast.                                                                                                                                                                                                 & Prompting function                                                                        \\
Rao et al.~\cite{rao2022denseclip}                   & Dense prediction                                                                                     & Discrete       & ADE20K                                                                                                                                                                                                                                                    & Language-guided fine-tuning with contextaware prompting                                   \\
Yao et al.~\cite{yao2023context}                   &  Prompt probing                                                                                                   & Manual         & LMC-EHRs                                                                                                                                                                                                                                                  & On BERT, RoBERTa, BioBERT, ClinicalBERT, 3 kinds of BioLMs, and 3 kinds of BlueBERTs.    \\
Zhang et al.~\cite{zhang2023gpt4mia}                  & Medical image classification, Transductive inference                                                  & Manual         & RetinaMNIST, FractureMNIST3D datasets                                                                                                                                                                                        & GPT4MIA                                                                                   \\
Yang et al.~\cite{yang2023bliam}                    & Drug synergy classification                                                                          & Discrete       &                                                                                                                                                                                                                                         & BLIAM                                                                                     \\
Sung et al.~\cite{sung2021can}                    & Predict/retrieve biomedical knowledge                                                                                                     & Manual         & CTD, UMLS, Wikidata                                                                                                                                                              & BioLAMA                                                                                   \\ \bottomrule
\end{tabular}
\label{classification}
\end{table*}

\subsubsection{Medical image analysis}
GPT4MIA utilizes the Generative Pre-trained Transformer (GPT) as a plug-and-play inference model for medical image analysis (MIA) classification tasks \cite{zhang2023gpt4mia}. The authors theoretically justify using GPT-3, a large pre-trained language model, for MIA. They develop prompt engineering techniques like improved prompt structure design, sample selection, and prompt ordering to enhance GPT4MIA's efficiency and effectiveness in detecting prediction errors and improving accuracy for image classification. GPT4MIA, working with established vision models, shows strong performance on these tasks.

\subsubsection{Clinical text classification}
HealthPrompt proposes a novel prompt-based clinical NLP framework that applies prompt engineering to pre-trained language models for classifying clinical texts without training data \cite{sivarajkumar2022healthprompt}. Deep learning models need large annotated datasets, but these are lacking for clinical NLP. HealthPrompt addresses this by using prompt-based learning to tune pre-trained language models for new tasks by defining prompt templates instead of fine-tuning the models. An analysis of HealthPrompt using six pre-trained language models in a no-data setting shows that prompts effectively capture the context in clinical texts and achieve good performance without training data.
\subsubsection{Mental health classification}
ChatGPT, an LLM-based model, demonstrates strong zero-shot performance in classifying social media posts for three mental health tasks: stress detection, depression detection, and suicidality detection \cite{lamichhane2023evaluation}. LLMs show promise for NLP mental health applications but need data, which ChatGPT addresses using prompt-based learning. ChatGPT outperforms the baseline model, indicating the potential of language models for mental health classification, especially when data is limited.

\begin{table*}[h]
\caption{Applications of prompt engineering in generation tasks}
\centering
\begin{tabular}{p{2.5cm}p{3.5cm}p{1.5cm}p{4cm}p{4cm}}
\toprule
\textbf{reference} & \textbf{task}                                                                    & \textbf{prompt type} & \textbf{dataset}                                                                                                                                                                                                         & \textbf{highlight}                                              \\ \midrule
Wang et al.~\cite{wang2023chatcad}                   & Medical text generation                                          & Manual         & MIMIC-CXR                                                                                                                                                                                                                & ChatGPT                                                         \\
Chambon et al.~\cite{chambon2022adapting}                   &Generate medical images & Discrete       & CheXpert, MIMIC-CXR                                                                                                                                                                                                      & Stable Diffusion model                                          \\
Yang et al.~\cite{yang2022prompting}                   & Medical text generation                                                                                 & Discrete       & MR, IMDB, SNLI                                                                                                                                                         & BiLSTM and BERT                                                 \\
Lu et al.~\cite{lu2023napss}                   & Automated text simplification                                                    & Discrete       & Cochrane paragraph-level medical text simplification dataset                                                                                                                                                             & BERT                                                            \\
Lehman et al.~\cite{lehman2022learning}                   & Medical question generating                                                      &                      & DiSCQ, MIMIC-III                       &                                                                 \\
Zuccon et al.~\cite{zuccon2023dr}                   & Medical text generation                                                    & Manual         & TREC 2021 and 2022                                                                                                                                                                                                       & ChatGPT                                                         \\
Hertz et al.~\cite{hertz2022prompt}                   & Images generation                                                                & Manual         &                                                                                                                                                                                                                       & Prompt-to prompt editing framework                              \\
Liu et al.~\cite{liu2022dynamic}                   &                                                                             & Discrete       & ACE,ERE                                                                                                                                                  &                                                                 \\
Abaho et al.~\cite{abaho2022position}                   & Health outcome generation                                        & Discrete       & EBM-NLP, EBM-COMET                                                                                                                                                                                                       & Position-attention mechanism                                    \\
Gao et al.~\cite{gao2023prefixmol}                   & Generating molecules                                                             & Continuous     & CrossDocked                                                                                                                                                                                                              & PrefixMol                                                       \\
Nori et al.~\cite{nori2023capabilities}                  & Medical text generation                                                       & Manual         & USMLE Sample Exam, USMLE Self Assessments, MedQA, PubMedQA, MedMCQA, MMLU                                                                                                                                             & GPT-4                                                           \\
Kasai et al.~\cite{kasai2023evaluating}                   & Medical text generation                                                                                 & Manual         & IGAKUQA dataset                                                                                                                                                                                                          & ChatGPT, GPT-3, and GPT-4                                       \\
Holmes et al.~\cite{holmes2023evaluating}                   &  Medical text generation                                                   & Manual         & 100-question multiple-choice                                                                                                                                                                                             & On ChatGPT (GPT-3.5), ChatGPT (GPT-4), Bard (LaMDA), and BLOOMZ \\
Chuang et al.~\cite{chuang2023spec}                   & Clinical note generation                                         & Continuous     & MIMIC-CXR                                                                                                                                                                                                                &                                                                 \\
Wang et al.~\cite{wang2022qrelscore}                   & Question generation                       &                      & SQuADv1, HotpotQA                                                                                                                                                                                                     &                                                                 \\
Du et al.~\cite{du2020schema}                   & Natural language generation                                                      & Discrete       & DSTC8                                                                                                                                                                                                                    &                                                                 \\
Jang et al.~\cite{jang2023exploring}                   & Medical note generation                                                 & Manual         & Korean National Licensing Examination                                                                                                                                                                                    & ChatGPT3.5 and ChatGPT4                                         \\
Chen et al.~\cite{chen2023towards}                   &  Generation                              & Continuous     & ROCO MedICaT MIMIC-CXR                                                                                                                                                                                                   & Vision-Language Pre-trained mode                                \\
Lyu et al.~\cite{lyu2023translating}                   & Clinical note generation                                                         & Manual         & Atrium Health Wake Forest Baptist clinical database.                                                                                                                                                                     & ChatGPT; GPT-4                                                  \\
Yan et al.~\cite{yan2022clip}                   & Entity clustering and entity set expansion                                       & Discrete       & eCoNLL2003, BC5CDR, WNUT 2017, WIKI                      & CLIP                                                            \\
Le et al.~\cite{le2022few}                   & Text generation                            & Discrete       & CHEMU-REF                                                                                                                                                                                                                & GPT-J-6B                                                        \\
Peng et al.~\cite{pengclinical}                   & Clinical concept extraction and relation extraction                              & Manual         & The 2018 National NLP Clinical Challenges and the 2022 n2c2 challenges &                                                                \\
\bottomrule
\end{tabular}
\label{generation}
\end{table*}

\subsection{Generation Task}

\subsubsection{Medical image generation}
Chambon et al. demonstrate how selective fine-tuning and meaningful evaluation enable the Stable Diffusion generative model to generate medical images from clinical prompts \cite{chambon2022adapting}. By customizing Stable Diffusion for the medical domain through focusing fine-tuning and assessing performance with clinically tailored metrics, the authors translate domain knowledge into model capabilities for sensitive generation where data is scarce. Their methods and results highlight the promise of prompt engineering to impart specialized expertise to models for nuanced generation to address real-world challenges. However, fully realizing this potential will require continued progress in techniques for model adaptation and domain-specific evaluation.

\subsubsection{Medical text generation}
NapSS presents a ``summarize-then-simplify" strategy to coherently simplify medical text \cite{lu2023napss}. NapSS generates summaries and narrative prompts to respectively train a model to extract key content and guide a generator to clarify it while preserving the flow of the text. Evaluated on medical text, NapSS improves on baselines in lexical, semantic, and human metrics. This highlights how multi-stage prompt engineering achieves nuanced, domain-relevant generation requiring sophisticated evaluation.
\subsubsection{Medical report translation}
Lyu et al. examine the LLM ChatGPT's performance in translating radiology reports into plain language for education \cite{lyu2023translating}. Evaluated by radiologists, ChatGPT shows promise but some inconsistency improvements with better prompts. ChatGPT generates both general and specific recommendations based on reports. Compared to GPT-4, which is a newer model, ChatGPT's performance is significantly outperformed, indicating LLMs’ potential for clinical use but requiring advanced models and prompts for optimized performance. 

Further research and applications of prompt engineering in generation tasks are shown in Table \ref{generation}.

\begin{table*}[h]
\caption{Applications of prompt engineering in detection tasks}
\centering
\begin{tabular}{p{2.5cm}p{3.5cm}p{1.5cm}p{4cm}p{4cm}}
\toprule
\textbf{reference} & \textbf{task}                                                 & \textbf{prompt type} & \textbf{dataset}                                                                                                                                                                                                   & \textbf{highlight}                           \\ \midrule
Liu et al.~\cite{liu2023deid}                   & Privacy protection task                                       & Manual         & The i2b2/UTHealth Challenge                                                                                                                                                                                        & ChatGPT and GPT-4                            \\
Feng et al.~\cite{feng2022beyond}                   & Multimodal knowledge learning                                 & Discrete       & MS-COCO, OpenImages Challenge 2019                                                                                                                                                                              & Multimodal knowledge learning         \\
Zhou et al.~\cite{zhou2022multi}                   & Event argument extraction                                     & Discrete       & ACE2005, RAMS, WiKiEvents                                                                                                                                                                                          &                                              \\
Yao et al.~\cite{yao2023automated}                   & Eviction status classification                                & Discrete       & HER notes                                                                                                                                                                                                          &                                              \\
Ruan et al.~\cite{ruan2023medical}                   & Medical intervention duration estimation                      & Discrete       & PASA Dataset, MIMIC-III                                                                                                                                                                                 & Language-enhanced transformer-based framework \\
Qin et al.~\cite{qin2022medical}                   & Medical image analysis                 & Discrete       & ISIC 2016, DFUC 2020, CVC-300, CVC-ClinicDB, CVC-ColonDB, Kvasir, ETIS, BCCD, CPM-17, TBX11k, Luna16, ADNI, TN3k                                                                                                   &                                              \\
Xing et al.~\cite{xing2023dual}                   & Object detection                                              & Discrete       & EuroSAT, Caltech101, OxfordFlowers, Food101, FGVCAircraft, DTD, OxfordPets, StanfordCars, ImageNet1K, Sun397, and UCF101 & Dual-modality Prompt Tuning (DPT) paradigm   \\
Ding et al.~\cite{ding2023exploring}                   & Multi Label Recognition               & Continuous     & MS-COCO, PASCAL VOC, NUSWIDE                   & CLIP                             \\
Lin et al.~\cite{lin2023exploring}                   & Multiple instance learning & Continuous     & Camelyon16, TCGA-NSCLC                                                                                                                                                                                             &                                             \\ \bottomrule
\end{tabular}
\label{detection}
\end{table*}

\subsection{Detection Task}

\subsubsection{Medical image detection}
Qin et. al show how prompt engineering enables vision-language models (VLM) pretrained on natural images to transfer knowledge to medical images \cite{qin2022medical}. Manual prompts containing expert medical knowledge improve VLM performance on medical detection tasks with no medical training data. Automatic prompts inject image details, further enhancing performance. Proposed approaches outperform default prompts on 13 medical datasets. Fine-tuned models beat supervised baselines, demonstrating VLMs can learn from limited medical data by transferring knowledge from natural images. This work establishes prompt engineering as a key to applying VLMs in medicine and paves the way for developing VLMs tailored to healthcare applications. 

\subsubsection{Medical text detection}
DeID-GPT is the first framework to de-identify text-free medical data using GPT-4's state-of-the-art NLP capabilities, which achieves the highest accuracy in removing private information while preserving its original meaning \cite{liu2023deid}. It requires no changes for different data types thanks to GPT-4's scale and in-context learning. DeID-GPT uses prompts to generate optimal de-identification results with little human input, showing the potential of ChatGPT and GPT-4 for automated medical text processing. It contributes a novel and highly effective approach to the important problem - enabling the use of medical text data while protecting patient privacy. DeID-GPT establishes GPT-4 and similar LLMs as a means to overcome grand challenges in healthcare through prompt engineering and natural language understanding.

Further research and applications of prompt engineering in detection tasks are detailed in Table \ref{detection}.

\begin{table*}[h]
\caption{Applications of prompt engineering in augmentation tasks}
\centering
\begin{tabular}{p{2.5cm}p{3.5cm}p{1.5cm}p{4cm}p{4cm}}
\toprule
\textbf{Reference} & \textbf{Task}                                                                                                                 & \textbf{Prompt type} & \textbf{Dataset}                                                                                                                                                               & \textbf{Highlight}                         \\ \midrule
Dai et al. \cite{dai2023chataug}                   & Medical text augmentation                                                                    & Manual         & PubMed20k Dataset                                                                                                                                                              & ChatGPT                                    \\
Milecki et al. \cite{milecki2023medimp}                   & Medical text augmentation    & Discrete      & DCE MRI, ID-RCB                                                                                                   & ChatGPT                                    \\
Lo et al. \cite{lo2020effective}                   & Mispronunciation detection                                                                                                    & Discrete       & MAS                                                                                                                                        &                                            \\
Yuan et al. \cite{yuan2023llm}                   & Privacy-Aware Data Augmentation                                                                                               & Manual         & Clinical Trial Data, Patient EHR Data                                                                                                                                          & LLM-based patient-trial matching \\
Li et al. \cite{li2023semantic}                   & Image augmentation                                                                                                            & Manual         & CIFAR-10, CIFAR-100, Caltech10, Stanford Cars, Flowers102, OxfordPets,DTD &                                            \\
Yang et al. \cite{yang2022seqzero}                   & Few-shot semantic parsing                                                                                                     & Discrete       & GeoQuery and EcommerceQuery datasets                                                                                                                                           & Filling in sequential prompts with LMs    \\ \bottomrule
\end{tabular}
\label{Augmentation}
\end{table*}

\subsection{Augmentation Task}

\subsubsection{Text data augmentation}
Dai et. al propose a new text data augmentation method called ChatAug \cite{dai2023chataug}. ChatAug uses the ChatGPT language model to rephrase sentences in the training data into multiple conceptually similar but semantically different samples, enlarging the sample corpus. They apply ChatAug to few-shot learning text classification in the medical domain and show improved performance over state-of-the-art data augmentation methods in terms of both testing accuracy and distribution of the generated samples.

\subsubsection{Transform clinical data into prompts}
Milecki et. al propose a new mutil-modal learning model called MEDIMP - Medical Images and Prompts \cite{milecki2023medimp}. MEDIMP translates clinical biomedical data into text prompts which as input of LLMs to produce augmented data. Contrastive learning and image-text pairs are then used to learn meaningful representations of medical images, i.e., renal transplant DCE MRI images. MEDIMP generates prompts using predefined sentence templates and ChatGPT, aiming to learn useful representations for prognosis of patient status 2-4 years after transplant. 

Further research and applications of prompt engineering in augmentation tasks are shown in Table \ref{Augmentation}.

\begin{table*}[h]
\caption{Applications of prompt engineering in QA tasks}
\centering
\begin{tabular}{p{2.5cm}p{3.5cm}p{1.5cm}p{4cm}p{4cm}}
\toprule
\textbf{Reference} & \textbf{Task}                                   & \textbf{Prompt type} & \textbf{Dataset}                            & \textbf{Highlight}                                           \\ \midrule
Liu et al. \cite{liu2022declaration}                   & Visual question answering                       & Discrete      & GQA                                         &                                                              \\
Liang et al. \cite{liang2022modular}                   & Visual question answering    & Continuous prompt    & VQAv2                                       & PromptFuse and BlindPrompt                                   \\
Kasai et al. \cite{kasai2023evaluating}                   & Answering medical questions                                 & Manual       & IGAKUQA dataset                             & ChatGPT, GPT-3, and GPT-4                                    \\
Holmes et al. \cite{holmes2023evaluating}                   &Answering medical questions                      & Manual       & 100-question multiple-choice                & ChatGPT (GPT-3.5), ChatGPT (GPT-4), Bard (LaMDA), and BLOOMZ \\
Liu et al. \cite{liu2022qaner}                   & Question Answer& Discrete      & MIT Movie, MIT Restaurant, CoNLL03  &                                                              \\
Jang et al. \cite{jang2023exploring}                   &  Answering medical questions                          & Discrete     & Korean National Licensing Examination       & ChatGPT3.5, ChatGPT4                                      \\
Li et al. \cite{li2022towards}                   &    Medical Instructional Video                  & Continuous    & MedVidQA  &                                                              \\
Lee et al. \cite{lee2023clinical}                   &  Answering medical questions                      & Manual      & Diabetes EHR dataset                        & Clinical Decision Transformer                 \\ \bottomrule              
\end{tabular}
\label{Question-Answering}
\end{table*}

\subsection{Question Answering Task}

\subsubsection{Medical text-based QA}
Singhal et. al present medical QA datasets, MultiMedQA and HealthSearchQA, and propose a framework for evaluating clinical LLMs along dimensions like factuality, precision, harm and bias \cite{singhal2022large}. To adapt LLMs like PaLM and FlanPaLM for the medical domain, the authors apply prompt engineering techniques like instruction prompt tuning, a parameter-efficient prompting approach for aligning LLMs to specialized domains using a few examples. The resulting model, Med-PaLM, shows encouraging improvements but still lags clinicians. This work highlights the importance of medical QA datasets and human-centered evaluation in creating beneficial clinical language models. By systematically evaluating tuned language models, the authors reveal gaps in comprehending medical knowledge and reasoning that must be addressed to develop models for healthcare.

\subsubsection{Medical image-based QA}
The open-ended medical visual QA work treats the task as a generative process. Sonsbeek et. al develop a network to map visual features to tokens that, alone with the question, directly prompt a PLM \cite{van2023open}. Exploring PLM fine-tuning strategies, their approach generates open-ended responses that outperform others on medical QA benchmarks like Slake, OVQA and PathVQA. This enables a PLM to understand medical images for open-ended medical visual QA tasks where answers are not constrained to a predefined set. The work points to promising directions for open-domain medical visual question answering using prompt-based generative models.

\subsubsection{Medical video-based QA}
Visual-Prompt Text Span Localization (VPTSL) proposes a novel approach to the temporal answering grounding in video (TAGV) task by formulating it as predicting the span of timestamped subtitles that matches the visual answer \cite{li2022towards}. To bridge the semantic gap between textual questions and visual answers, VPTSL introduces visual prompts—highlight features obtained from video-text highlighting using the question. These visual prompts are fed into the PLM along with the subtitles and question. A text span predictor then models these visual and textual prompts to predict the subtitle span. VPTSL outperforms the state-of-the-art on MedVidQA by a large margin (28.36\% in mIOU), showing the effectiveness of visual prompts and the text span predictor for medical video QA. This work enables PLMs to ground temporal answers in instructional videos by prompting them with contextual visual information.

Further research and applications of prompt engineering in question answering tasks are detailed in Table \ref{Question-Answering}.

\begin{table*}[b]
\caption{Applications of prompt engineering in inference tasks}
\centering
\begin{tabular}{p{2.5cm}p{3.5cm}p{1.5cm}p{4cm}p{4cm}}
\toprule
\textbf{Reference}                                       & \textbf{Task}                                                              & \textbf{Prompt type} & \textbf{Dataset}                                                                                                                              & \textbf{Highlight}                             \\ \midrule
Wang et al. \cite{wang2022prompt}                                                         & Biomedical natural language inference                                      & Discrete       & MedNLI, MedSTS                                                                                                                                &                                                \\
Long et al. \cite{long2023can}                                                         & Building causal graphs                                                     & Manual         &      & GPT-3                                          \\
Li{\'e}vin et al. \cite{lievin2022can}                                                         & Medical natural language inference                                                                           & Manual         & USMLE, MedMCQA, PubMedQA                                                                                                                      & GPT-3.5                                        \\
Kim et al. \cite{kim2022leveraging}                                                         & Natural language interaction  & Discrete       & Extractive and multiple-choice QA                                                                                                             & GPT-3                                          \\
Li et al. \cite{li2022advance}          & Arithmetic Reasoning, Commonsense Reasoning, Inductive Reasoning           & Discrete       & GSM8K AsDiv, MultiArith, SVAMP, SingleEq, CommonsenseQA, StrategyQA, CLUTRR   & Davinci, text-davinci-002 and codedavinci-002 \\
Gao et al. \cite{gao2022pal} & Mathematical problems, Symbolic reasoning, Algorithmic problems            & Discrete       & GSM8K, SVAMP, ASDIV, MAWPS, BIG-Bench Hard   & ProgramAided Language models                  \\ \bottomrule
\end{tabular}
\label{Inference}
\end{table*}

\subsection{Inference Task}

\subsubsection{Natural language inference in radiology}
Wu et. al evaluate ChatGPT and GPT-4 on a radiology natural language inference task \cite{wu2023exploring}. They implement zero-shot and few-shot prompts in the models. The zero-shot prompt provides only task instructions and sentence-question pairs, requiring the models to determine entailment with no labeled examples. The few-shot prompt incorporates 10 labeled sentence-question examples to provide in-context learning before the evaluation pair. By designing these prompts with varying levels of contextual information, the authors show that ChatGPT and GPT-4 can achieve over 50\% accuracy on the radiology task without requiring large amounts of training data. GPT-4 outperforms ChatGPT, indicating its greater capability. The results demonstrate the feasibility of building generic models that can perform well across different domains.

\subsubsection{Causal reasoning about medical variables}
Long et. al examine whether GPT-3 can accurately predict the presence or absence of edges between variables in causal graphs based on medical context \cite{long2023can}. They evaluate GPT-3 using different prompts (e.g. declarative v.s. interrogative) and linking verbs (e.g. ``causes" v.s. ``correlates with"). They find GPT-3's performance varies based on these factors, indicating its sensitivity to user input. With well-designed prompts using ``causes", GPT-3 achieves over 50\% accuracy on the tested graphs. Further, the authors select 3 causal graphs of different complexities from the medical literature. For each graph, they generate 2000 randomly permuted sentence pairs from the graph. GPT-3 predicts whether a causal edge exists between each pair. Its accuracy is the highest (70\%-85\%) on the simplest graph. Though performance decreases for the complex graphs, it remains well above the 50\% random baselines, demonstrating GPT-3's potential for reasoning about medical variables.

Further research and applications of prompt engineering in inference tasks are shown in Table \ref{Inference}.

\subsection{A Recap of Prompt Engineering for Classification, Generation, Detection, Augmentation, QA, and Inference}
These studies highlight the potential of using prompt engineering for pre-trained language models to advance clinical NLP and a range of medical AI applications, especially when annotated data is scarce. Prompt-based learning tunes models for new tasks by defining task templates instead of fine-tuning, helping models achieve strong performance in zero-shot learning scenarios where they generalize to new classes without examples. Large pre-trained language models provide an ideal initialization for prompt engineering across medical domains and tasks when combined with effective prompt design. 

Though more work is needed, these papers demonstrate the promise of prompt engineering and LLMs for medical AI with limited data. Prompt-based learning could help address lacks of annotated data and advance applications like classification, generation, detection, enhancement, question answering, and reasoning by unlocking models' capabilities through tailored prompts. Strong zero-shot performance suggests prompt engineering may reduce data needs for medical AI versus traditional supervised learning.

\section{Challenges and Future Directions}

This chapter discusses the challenges, current research directions, and future opportunities and development directions of prompt-based methods.

Firstly, the challenges in prompt engineering are addressed, including the data scarcity~\cite{azaria2016medrec} in the medical NLP domain, the interpretability of models, and inherent issues in prompt engineering.

Secondly, the current research directions are introduced, which include prompt generation~\cite{he2022hyperprompt,liu2023pre}, prompt optimization~\cite{ma2023impressiongpt}, mutil-modal data processing~\cite{wang2023chatcad}, and deep reinforcement learning~\cite{franccois2018introduction}. These research directions aim to improve the effectiveness and applicability of prompt-based methods, further promoting the development of the NLP field.

Lastly, the opportunities and future development directions of Prompt-based methods are explored, such as the development of multitask and mutil-modal processing and the expansion of innovative Prompt design and intelligent Prompt generation~\cite{ma2023impressiongpt,white2023prompt}. These opportunities and development directions provide a broad space and prospects for the future development of Prompt-based methods.

\subsection{Challenges in Prompt Engineering}
When it comes to NLP tasks in the medical field, prompt engineering faces several challenges. First, medical data is usually limited and specialized, making it difficult to cover all domain knowledge and thus restricting the model's generalization ability~\cite{liu2023deid,wang2023chatcad}. Secondly, the medical field is rich in terminology and domain knowledge, but often complex, which requires effective integration of this knowledge into prompts~\cite{zhou2007temporal}. Additionally, different medical tasks require different prompt designs, which requires balancing the complexity and interpretability of prompts while also utilizing existing medical domain knowledge to guide the model to generate high-quality predictions~\cite{singhal2022large}. Finally, the prompt design also needs to consider how to avoid introducing any human bias or erroneous information to ensure the fairness and accuracy of the model~\cite{crothers2022machine}.

It should be noted that NLP tasks in the medical field are highly challenging as they involve a large amount of domain knowledge and specialized terminology, which may not be easily understood or processed by general NLP models~\cite{chen2023see}. Therefore, the importance of prompt engineering in the medical field is self-evident. These challenges include, but are not limited to:

Data scarcity: In the medical field, many tasks require the use of specific medical data, which is often very scarce and difficult to obtain~\cite{liu2023deid,yao2022extracting}. Moreover, the special nature of medical data involves ethical issues, which poses a great challenge to prompt engineering. The lack of sufficient data makes it difficult to design accurate and effective prompts.

Data uncertainty: The medical field involves numerous domain knowledge and terminology, which may have different interpretations and usage in different texts, leading to increased uncertainty in prompt design~\cite{yao2022extracting,zuccon2023dr}.

Model interpretability: In the medical field, model interpretability is particularly important. As it involves human health and life, the model's prediction results need to be reasonably explained and justified. Therefore, in the medical field, prompt design not only considers the accuracy of the model but also its interpretability~\cite{liu2023deid}.

Additionally, prompt engineering in the medical field faces common challenges in other fields, such as prompt engineering's self-consistency issues~\cite{wang2023experimental}, prompt leakage~\cite{perezignore}, and adversarial issues~\cite{mcguffie2020radicalization}, among others. These challenges in medical tasks are more severe than in other fields. Future research needs to fully consider these challenges and propose corresponding solutions.

In summary, prompt engineering in the medical field is critical to the success of NLP tasks, but it also faces various challenges, such as data scarcity\cite{rajpurkar2022ai,liu2022federated} and uncertainty\cite{wang2022semi}, model interpretability\cite{barragan2022towards,li2022interpretable}, self-consistency\cite{wang2022self}, and adversarial issues\cite{rasool2022security}. Addressing these challenges requires innovative solutions and a comprehensive understanding of the medical domain knowledge and its related terminology.

\subsection{Current Research Directions}
As the development and application of artificial general intelligence (AGI) models continue to advance, researchers in the medical field have also begun to apply them to various medical tasks. Currently, the research direction of prompt engineering in the medical field is very diverse, covering different types of prompt designs~\cite{wang2023chatcad}, multi-modal data processing~\cite{lin2023exploring}, and deep reinforcement learning~\cite{Zhong2023ChatABLAL}, among other aspects.

In terms of prompt design methods, it is necessary to consider the specific characteristics of the task and data features to make appropriate choices. These include simple manually designed template-style prompts, prompts based on knowledge graphs\cite{andrus2022enhanced}, prompts generated based on natural language\cite{gilson2023does}, and prompts that can be embedded as trainable parameters in the model for automatic generation\cite{niu2023ct}. In multi-modal data processing, integrating text and image data requires the design of appropriate prompts to guide the model to handle and analyze different types of data, thereby improving the model's performance and interpretability\cite{behrad2022overview}. Additionally, deep reinforcement\cite{jacob2022autonomous} learning can continuously optimize prompt design through autonomous learning, further improving the performance of the model.

In summary, the current research direction of prompt engineering in the medical field is diverse, and researchers can choose appropriate methods and techniques based on the requirements of the task and data characteristics to improve the performance and interpretability of the model.

\subsection{Opportunities and Future Directions }

In the field of medicine, prompt engineering presents a wide range of opportunities and future directions for application. The development of LLMs such as ChatGPT and GPT-4 provides tremendous potential for prompt engineering in the medical field~\cite{liu2023summary}. As an essential part of prompt-based technology, prompt engineering can assist large models in understanding and processing medical data more accurately~\cite{chen2022applications}, from manual prompts to automated prompts, improving the automation and intelligence of prompts~\cite{thakur2023automated}. Applying automated algorithms to prompt engineering, utilizing data-driven methods to generate more intelligent and efficient prompts, can enhance the efficiency and accuracy of NLP tasks in the medical field~\cite{martin2023hypertension}. With the emergence of mutil-modal information in the medical field, combining multiple modalities such as text, images, and speech can better address practical problems in the medical domain\cite{azam2022review}. Additionally, conducting multi-task research by integrating multiple medical research tasks can provide better services for clinical healthcare\cite{chengoden2023metaverse,buchlak2022natural}.  In conclusion, prompt engineering is a promising area for medical NLP research, and future studies should focus on developing more intelligent and automated prompt engineering methods and exploring the combination of multiple modalities and tasks.

\section{Conclusion}
In this section, we will provide a summary of the review and discuss its limitations and contributions.

\subsection{Summary of the Review}
Overall, this review article provides a comprehensive overview of the different prompt engineering methods and challenges in the context of NLP tasks in the medical field. We have presented different prompt design methods, including manual and automated, discrete and continuous, and provided examples to illustrate their practical applications in medical settings. Additionally, we have discussed the different prompt engineering methods used for various medical tasks, such as classification, generation, detection, argumentation, reconstruction, question answering, prediction, and inference tasks.

One key finding of this review is that prompt engineering is a promising approach to improving the performance of NLP tasks in the medical field. By carefully designing prompts that are tailored to specific tasks and domains, researchers can achieve significant improvements in accuracy and efficiency.

However, there are also several challenges and limitations to prompt engineering that need to be addressed in future research. Moreover, the effectiveness of prompts may depend on the specific characteristics of the task and the domain, which can vary widely across different medical applications.

\subsection{Limitations of the Review}

Despite the comprehensive literature search process and the inclusion of 333 relevant studies, it is possible that some relevant research in the field of medical NLP prompt engineering may have been missed. Therefore, the scope of this review may not be entirely exhaustive, and there could be other important studies that are not covered.

Additionally, although the article covers a range of medical tasks and the prompt engineering methods employed in each of them, there may be other tasks or research methodologies that are not explored. Future research in this field should, therefore, continue to investigate the potential of prompt engineering in other areas of medical NLP and explore additional methods that may not have been considered in this review.
\subsection{Conclusion and Contributions}
This review article makes a significant contribution to the field by providing a comprehensive and systematic overview of prompt engineering methods for NLP tasks in the medical domain. By covering a wide range of methods from manual to automated, and from discrete to continuous, the article serves as a valuable reference and guide for researchers in the medical NLP field to help them choose and design prompts that can improve model performance and efficiency. Additionally, the article discusses the unique challenges posed by medical NLP tasks and explores the future research directions for prompt engineering in this field. Overall, this review article is a valuable resource for medical NLP researchers and provides insightful guidance for future research in this rapidly growing field.

\subsection{Recommendations for Future Research}

In summary, the future of NLP in the medical field looks promising, with more advanced techniques such as prompt engineering and multi-modal integration being explored. We can expect to see more research in applying state-of-the-art models like ChatGPT and GPT-4 to medical NLP tasks, as well as developing more efficient and accurate prompt engineering methods\cite{kutela2023chatgpt}. Furthermore, there is a need to explore other areas of NLP in medical filed\cite{wu2023exploring}, such as text-based medical image analysis and building medical knowledge graphs.

As the field continues to evolve, the development of large models such as GPT-4 will further advance NLP research. However, this will also require innovation and breakthroughs in both hardware and software to accommodate the higher computing and resource demands.

Overall, future work in this field will require a comprehensive and precise approach that incorporates multi-modal information and large model technology to advance medical NLP research and applications. With the potential for cross-disciplinary collaboration between medical professionals and NLP researchers, the possibilities for improving healthcare through NLP are vast.

\section{Acknowledgment}

This work was supported by Faculty Construction Project under Grants No. 22SH0201178, 23SH0201228; Foundational Research in Specialized Disciplines under Grant No. G2023WD0146; National Natural Science Foundation of China under Grants No. 62276050; National Natural Science Foundation of China under Grants No. 62001410; National Natural Science Foundation of China under Grants No. 61976131.



\bibliographystyle{IEEEtran.bst}
\bibliography{mybib}

\end{document}